\documentclass{bmcart}

\usepackage[utf8]{inputenc} 

\def\includegraphics{}

\startlocaldefs
\endlocaldefs

\begin{document}

\begin{frontmatter}

\begin{fmbox}
\dochead{Research}


\title{Literature Triage on Genomic Variation Publications by Knowledge-enhanced Multi-channel CNN}


\author[
   addressref={aff1},
   email={lvchenhui@seu.edu.cn}
]{\inits{JE}\fnm{Chenhui} \snm{Lv}}
\author[
   addressref={aff1},
   email={luqian@seu.edu.cn}
]{\inits{JRS}\fnm{Qian} \snm{Lu}}
\author[
   addressref={aff2},                   
   corref={aff1,aff2},                    
   email={x.zhang@seu.edu.cn}   
]{\inits{JE}\fnm{Xiang} \snm{Zhang}}

\address[id=aff1]{%
  \orgname{College of Software Engineering, Southeast University},
  \city{Suzhou},
  \cny{China}
}
\address[id=aff2]{
  \orgname{School of Computer Science and Engineering, Southeast University},
  \city{Nanjing},                              
  \cny{China}                                    
}


\end{fmbox}


\begin{abstractbox}

\begin{abstract} 
\parttitle{Background} 
To investigate the correlation between genomic variation and certain diseases or phenotypes, the fundamental task is to screen out the concerning publications from massive literature, which is called literature triage. Some knowledge bases, including UniProtKB/Swiss-Prot and NHGRI-EBI GWAS Catalog are created for collecting concerning publications. These publications are manually curated by experts, which is time-consuming. Moreover, the manual curation of information from literature is not scalable due to the rapidly increasing amount of publications. In order to cut down the cost of literature triage, machine-learning models were adopted to automatically identify biomedical publications.

\parttitle{Methods} 
Comparing to previous studies utilizing machine-learning models for literature triage, we adopt a multi-channel convolutional network to utilize rich textual information and meanwhile bridge the semantic gaps from different corpora. In addition, knowledge embeddings learned from UMLS is also used to provide extra medical knowledge beyond textual features in the process of triage.

\parttitle{Results}
We demonstrate that our model outperforms the state-of-the-art models over 5 datasets with the help of knowledge embedding and multiple channels. Our model improves the accuracy of biomedical literature triage results.

\parttitle{Conclusions}
Multiple channels and knowledge embeddings enhance the performance of the CNN model in the task of biomedical literature triage.
\end{abstract}




\end{abstractbox}
%

\end{frontmatter}



\section*{Background}
One of the critical questions in biomedicine is to identify the contribution of our genomic variation plays in the chance of developing certain diseases or phenotypes. Although genomic variations are not the only decisive factor for certain diseases, scientists are increasingly using genetic knowledge to predict the risk of developing different illnesses\cite{tam2019benefits}. Naturally occurring genetic variants, both rare and common, can provide insight into disease mechanism and protein function. The research on genetic risks is entering a new era, where the results of large-scale DNA sequencing may be combined with other diagnostic indicators to define genetic risk. This may have a major impact on individuals’ lifestyle choices and direct personalized medicine to provide the best treatment for a patient’s genetic profile.

In this decade, there has been a dramatic increase in publications on the subject of identifying genomic variation. These publications provide an unprecedented opportunity to investigate the impact of common variants on complex diseases. However, identifying these publications can be challenging, and the vast wealth of data contained within these publications is inaccessible to researchers without systematic cataloging and summarization of the observed associations. Currently, the process of literature triage is still mainly conducted in a manual way. Expert-curated knowledge bases, such as UniProtKB/Swiss-Prot \cite{uniprot2016uniprot} and NHGRI-EBI GWAS Catalog \cite{buniello2018nhgri}, are vital resources for scientists to the search, visualization and exploitation of literature. But they require domain experts to collect and manually curate high-quality information from massive literature, which turns out to be a highly time consuming and costly process.

It has been studied that the manual curation of information from biomedical literature is not scalable due to the fast-growing number of publications \cite{baumgartner2007manual, schnoes2009annotation}. Text mining as a high-throughput computational technique can prominently save the human efforts on curation. Currently, databases integrate manual curation by domain experts with support from automated text-mining models \cite{winnenburg2008facts, muller2018textpresso, wei2017tmvar, singhal2016text, pletscher2015diseases, ferrari2015genome, lee2018scaling}. In this paper, we focus on the first task of the manual literature curation process, which is called literature triage \cite{poux2017expert}. This process filters among a long list of publications for only the documents that seem to be potential candidates for full curation. Machine learning models are playing an increasingly important role. Differ from the traditional manual triage by keyword-based query, scalable and accurate automatic triage speeds up the process by learning curated data and making predictions on the relevance of raw literature. 

Web-based curation support systems, such as PubTator \cite{poux2017expert} and Textpresso\cite{muller2018textpresso} help biocurators to search and annotate biomedical literature. Curators can read publications that are highlighted and pre-annotated by automated named-entity recognition tools. Curators can also easily upload and generate their curation collection and save their curation results. However, the process of literature triage in PubTator and Textpresso is highly relying on human judgment, instead of machine prediction. In the system of mycoSORT \cite{almeida2014machine}, three machine learning algorithms (Naïve Bayes, Support Vector Machine and Logistic Model Trees) are used in literature triage for PubMed abstracts in the mycoCLAP database \cite{murphy2011curation}. Since mycoSORT is focusing on the literature of enzyme families, the adopted feature extraction and selection approach is consequently domain-dependent, and cannot be naturally generalized into other databases. tmVar \cite{wei2013tmvar,wei2017tmvar} is another system that helps curators collecting genomic variant information from literature and integrate it with dbSNP \cite{sherry2001dbsnp} and ClinVar \cite{landrum2015clinvar}. tmVar detects mentions of genomic variants in literature and normalizes them into unique dbSNP RSIDs. Although tmVar may not facilitate literature triage directly, it is quite useful for positive or negative sampling in literature triage systems \cite{uniprot2016uniprot,lee2018scaling,chen2017document,hsu2018assisting,allot2018litvar}. The READBiomed system \cite{chen2017document} creates term lists describing interactions, mutations, and expected effects on interactions mutations, and it trains an SVM-based literature classification model using these term lists along with a range of standard bag-of-word features. In \cite{hsu2018assisting}, literature triage is achieved by different classifiers (Glmnet, SVM and CNN) for the task of Human Kinome Curation. However, the work still requires a time-consuming feature engineering, including the selection of frequency, location, and linguistic features, together with manually generated keyword groups of the genetic disease field. A CNN-based system proposed in \cite{lee2018scaling} provides a state-of-the-art model for the triage of genomic variation literature. This work is quite relevant to our work. It investigates three knowledge bases: UniProtKB/Swiss-Prot, NHGRI-EBI GWAS Catalog and mycoSORT, and it employs a deep learning model without a tedious process of feature engineering. Empowered by deep learning, this work greatly outperforms triage systems with traditional machine-learning models both in accuracy and generalization.

There is a prevalent trend that state-of-the-art triage systems start to explore deep learning models to identify genomic variation publications. However, existing works on literature triage have not fully exploited the recent development of deep learning paradigms. Triage on biomedical literature is not only a text-based mining task but also a knowledge-intensive task due to the domain-specific essence of the task. Bio-medical knowledge, especially genomic knowledge, should play a crucial role in predicting relevant genomic variation publications. However, this perspective has not been fully considered in recent works. The problem of knowledge deficiency in literature triage can be tackled with the paradigm of knowledge embedding \cite{wang2017knowledge}, which transforms explicit knowledge into computable and low-dimensional vectors that simplifies the manipulation while preserving the inherent structure of the knowledge. 

Besides, text features in current deep-learning-based triage systems are usually represented as low-dimensional pre-trained word vectors. Although the technology of word embedding has proved to be effective in a large number of learning tasks, a single-sourced or single-version word embedding is usually biased and is inefficient to integrally cover different aspects of meanings the text conveys, as stated in \cite{quan2016multichannel}. The bias of embedding may decrease the accuracy of triage. The problem of embedding bias can be solved by introducing multi-channel deep learning models, which takes multi-version word embeddings from diverse corpus as multi-channel input in the process of literature triage.

We hereby propose a Knowledge-enhanced Multi-channel CNN model (KMCNN in short) to accurately identify publications that are relevant to genomic variation. In KMCNN, the biomedical knowledge base UMLS \cite{tam2019benefits} is fully exploited to tackle the problem of knowledge deficiency. UMLS is an integration of comprehensive biomedical concepts and terms, presenting their meanings and hierarchical categories. The knowledge in UMLS is transformed into distributed knowledge vectors by a graph embedding model. Two versions of word embeddings are generated from PubMed and Wikipedia corpus separately. The KMCNN model takes each version of word embedding in a channel, in conjunction with knowledge vectors, to fulfill the literature triage task. We benchmark our model against the state-of-the-art triage system stated in \cite{lee2018scaling} on three datasets.

\section*{Methods}
We choose UniProtKB/Swiss-Prot, the NHGRI-EBI GWAS Catalog, and MycoSet as benchmark datasets in our experiment. In addition, we developed two extra datasets, named GWAS Catalog2019(a) and GWAS Catalog2019(b).

We propose the KMCNN model which features multiple channels and knowledge embedding. On one hand, multiple channels in CNN fill the vocabulary gap among different domains into consideration. On the other hand, textual features and medical knowledge features are jointly fused in literature triage so that biomedical literature triage could be improved by medical knowledge.

\subsection*{Dataset Selection}
In our experiment, we first use three datasets, including UniProtKB/Swiss-Prot, the NHGRI-EBI GWAS Catalog, and MycoSet as the benchmark datasets.

To compare our work with the state-of-the-art methods, we follow their synchronous train/test split, which means the training data and testing data are both sampled from the same time period. An obvious defect of synchronous split is that it is hard to measure the model’s capability on predicting the relevance of biomedical literature in the future. To address this problem, we develop two additional datasets based on GWAS Catalog, which is call GWAS Catalog 2019(a) and 2019(b).

In these two datasets, we adopt an asynchronous train/test split. Publications before 2018/1/1 (old publications) are put into the training set and the validating set, and the others (new publications) are put into the testing set. All positive samples are obtained directly from the website of GWAS catalog, including a total of 3,522 publications from 2008/6/16 to 2019/6/14. We use different negative-sampling strategies in each dataset. In GWAS Catalog 2019(a), 2,500 negative samples are randomly downloaded from PMC. In GWAS Catalog 2019(b), in order to create a negative sample collection that contain the most ambiguous samples rather than those absolutely irrelevant ones, 100,000 samples were firstly randomly downloaded from PMC. Then tmVar was utilized to extract articles that contains concepts denoting genes and diseases. On the basis of the practice which was introduced by \cite{lee2018scaling}, considering of some GWAS-related articles, e.g. some systematic reviews that involves GWAS researches, may also confuse the model although they might not mention specific genes or diseases, 18 keywords with the highest frequency in positive samples were also used for query in PMC. The results of the query and the extracted articles by tmVar were both added to the negative sample set after being de-duplicated with the 3,522 positive samples. Finally, ~4,000 medical literature remained as negative samples in GWAS Catalog 2019(b).

\subsection*{Multi-channel Word Embedding}

The problem of literature triage is usually solved by text classification. Literature classifiers are trained by exploiting text features in medical abstracts. In deep-learning-based triage models, such as \cite{lee2018scaling}, text features in medical abstracts are distributively represented as word vectors, which are pre-trained using large-scale corpus. In the past years, pre-trained word embeddings have been continuously developed from Word2Vec to Glove \cite{pennington2014glove} and FastText \cite{joulin2016bag}, which makes it possible to achieve satisfactory performance on specific downstream tasks with only slight fine-tuning efforts.

However, there are usually vocabulary and semantic gaps between different corpora, which lead to different versions of word representations. As stated in \cite{kim2014convolutional}, each version of representation shows merits in different tasks, and it is impossible to cover all corpus existed and to pre-train a comprehensive word representation for all tasks. A feasible choice to collect different versions of word representations and feed them to the classifier as multiple channels.

In this paper, we present a biomedical literature triage approach based on convolutional neural networks with both knowledge enhancement and multiple channels. As a variation of CNN, a multi-channel CNN \cite{kim2014convolutional} refers to a model with multiple sets of word embedding vectors. Each set of vectors is treated as a ‘channel’ and all the channels share common filters. For each filter, there are multiple vectors of the same size after convolution, and the average of them is taken into the convolution layer. The following pooling layer and the fully connected layer is the same as the counterpart in CNN.

Biomedical literature triage is a domain-specific task that involves tremendous medical terminologies. Therefore, word vectors pre-trained on biomedical corpora, e.g. PubMed, is a suitable choice for this task. Two sets of word embedding are adopted because one single set of word vector may bring biases and multiple sets of embedding could provide complementary information from different aspects.

In our model, the 200-dimensional word2vec vectors of Pyysalo et al. \cite{moen2013distributional} are used as one channel, which were pre-trained on all the PubMed abstracts and PubMed Central open access full texts. Another channel is Wikipedia-PubMed-and-PMC-W2V, which is a set of word vectors induced on a combination of PubMed and PMC texts with texts extracted from a recent English Wikipedia dump. 

\subsection*{Knowledge Embedding on UMLS}
The word vectors mentioned above could only capture the textual features in the medical literature. However, medical knowledge could hopefully offer more information to the model. To utilize medical knowledge, a practical way is to exploit medical knowledge bases, such as UMLS, and to learn the correlations between medical concepts from the knowledge bases. UMLS is an integration of comprehensive biomedical concepts and terms, presenting their meanings and hierarchical categories. It provides a semantic network that demonstrates broad categories and their relationships. Graph embedding approaches, including translational models \cite{bordes2013translating}, semantic matching models \cite{socher2013reasoning}, path-based graph embedding approaches like DeepWalk \cite{perozzi2014deepwalk} and Node2Vec \cite{grover2016node2vec} could capture the information in the semantic network. However, they could hardly deal with a multi-mode graph (a graph with different types of nodes) and utilize structural similarity in graphs.

We propose a UMLS2Vec model to overcome the above weaknesses, which learns a distributed representation of medical concepts in UMLS. In this model, the semantic network of UMLS is the input. Two random-walk strategies are proposed to capture the structural features for each medical concepts. One strategy is a homophily-oriented path-finding strategy (H-path for short), which focuses on capturing linkage adjacency between medical concepts, the other is a structural-equivalency-oriented path-finding strategy (S-path for short), which captures the structural similarities between medical concepts.

The input of UMLS2Vec is a knowledge graph $ g=\langle V, E, C\rangle$. V is the collection of nodes in $g$, which in our work represents the set of medical concepts in UMLS. E is the edge set of g representing the relationship between concepts. In UMLS, each concept is categorized into a semantic type. $C$ represents a total of 187 semantic types in UMLS. The output of UMLS2Vec is concept paths of random walks generated by H-path and S-path. We denote the output of Concept2Vec as $S = S^h \cup S^s$, where $S^h$ and $S^s$ are paths generated by H-path and S-path respectively.

In H-path strategy, $W_x = \langle x, W_x^1, W_x^2, \dots, W_x^l\rangle$ denotes a random walk rooted from concept $x$, where $l$ is the length of the path. It is a stochastic process with random variables $W_x^1, W_x^2, \dots, W_x^i$, such that $W_x^{i+1}$  is a concept randomly selected from the direct neighbors of the predecessor $W_x^i$. For each concept in UMLS, a number of random walk sequences are generated and put into $S^h$.

In S-path strategy, two concepts share similar patterns if they have the same types of neighboring concepts. Two concepts are structurally similar if they are close. Given a starting concept $v_i$ in the $|C|$-dimensional space, a virtual hypercube taking $v_i$ as the center can be defined, which has an edge length of $2r$ and is used for outlining the scope of structurally-close neighbors of $v_i$. Taking a concept x as the starting concept, a random neighbor in the square will be selected as the successor of x in an S-path. Iteration continues until a fix-length S-path is generated.

\subsection*{KMCNN Model for Literature Triage}
The architecture of KMCNN is shown in Figure \ref{kmcnn}.  Our model modifies the word embeddings in multiple channels and adds knowledge embedding based on the multi-channel CNN model developed by \cite{kim2014convolutional}. It is intuitive that multiple versions of word vectors from different channels could bring richer information and offset the biases one single channel may result in. The concatenation of multiple sets of word vectors and knowledge embedding could be utilized by the model to generate features and feature maps.

The representation for each publication is two $n\times k$ vectors, and k is the summation of $dw$ and $dk$. n is the fixed sequence length, which denotes the amount of words in an article. We do not take the entire contents of publication as the input text of models. The input text includes the title, abstract, PubMed id, journal’s name and publication type of each biomedical literature. $dw$ denotes the dimensions of the two sets of word vectors, both are 200. $dk$ denotes the dimension of the knowledge embeddings, which is 108.

 In the convolutional layer, there can be different settings of kernel sizes. For example, as Figure \ref{kmcnn} shows, if we take 1, 2, and 3 as the width of filter, there would be three sets of matrixes with the shape of $n\times1$, $(n-1) \times1$, and $(n-2) \times1$ respectively in the convolutional layer. The amount of each kind of kernel is M. We take the average of multiple sets of convolutional results coming from different channels. A max-pooling operation is applied.
 
  \begin{figure}[h!]
  \caption{\csentence{The architecture of KMCNN}}
  \label{kmcnn}
 \end{figure}

\section*{Results}
\subsection*{Data and Methods}

Table \ref{datasets} shows the statistics of the five datasets used in our experiments.

\begin{table}[h!]
\caption{Statistics of five datasets in the experiements}
      \begin{tabular}{cccc}
        \hline
        Datasets  &Training set size  &Testing set size   & Pos/ Neg\\ \hline
        UniProtKB/Swiss-Prot & 20,981 & 2,331 & 1: 1\\
        NHGRI-EBI GWAS Catalog & 4,657 & 500 & 1: 1\\
        MycoSet &3,020  & 335   & 1: 1\\
        GWAS Catalog 2019(a)  &10,926  & 2,855  & 3: 7\\
        GWAS Catalog 2019(b) &20,981 & 2,331   & 1: 1\\ \hline
      \end{tabular}
      \label{datasets}
\end{table}

In KMCNN, the detailed experimental settings are shown in Table \ref{settings}.  The filter sizes are set as 1, 2 and 3 respectively, which corresponds to 1-gram, 2-gram, and 3-gram language models. The drop rate is set to 0.5 in order to avoid over-fitting.

We compare our method against three baseline methods in our experiments. These baselines include one state-of-the-art methods and two ablation model of KMCNN.

Lee et al. proposed a CNN-based model for literature triage in genomic variation resources \cite{lee2018scaling}. Their model achieved better performance on UniProtKB/Swiss-Prot, NHGRI-EBI GWAS Catalog, and MycoSet comparing to other traditional machine learning methods. But their model did not adopt multiple channels or knowledge embedding. We take their model as one of our baseline methods.

\begin{table}[h!]
\caption{Experiment settings}
      \begin{tabular}{cc}
        \hline
        Parameters  &Values\\ \hline
        max sequence length & 1000\\
        embedding dimension & 200\\
        filter sizes &1,2,3\\
        number of filters  &2048\\
        hidden layer dimension &100\\
        drop rate  &0. 5\\
        learning rate  &1e-5\\
        epochs  &50\\
        hidden layer dimension  &30\\ \hline
      \end{tabular}
      \label{settings}
\end{table}

Besides the study of Lee et al., we also experiment on multi-channel CNN (without knowledge embeddings) short for MCNN and knowledge-enhanced CNN (without multiple channels) short for KCNN.

\subsection*{Evaluation on Biomedical Literature Triage}
We trained our model on different datasets, including UniProtKB/Swiss-Prot, NHGRI-EBI GWAS Catalog, MycoSet, and two datasets developed by ourselves, and compare the performance between it and other baseline methods.

The F1 score, precision, and recall of KMCNN and baseline methods are demonstrated respectively in Table \ref{f1}, Table \ref{precision}, and Table \ref{recall}. The parameters of the model are shown in Table \ref{settings}.

From Table \ref{f1}, Table \ref{precision}, and Table \ref{recall} we can tell that multi-channel CNN and knowledge-enhanced CNN both outperformed the deep-learning-based model proposed by Lee et al. Generally, KMCNN achieved the best performance on all of the four datasets in terms of F1 score, precision and recall.

\begin{table}[h!]
\caption{F1 score of KMCNN and other methods}
      \begin{tabular}{cccccc}
        \hline
          &UniProtKB/ & NHGRI-EBI   &MycoSet  &GWAS   &GWAS \\ 
          &Swiss-Prot & GWAS Catalog  & &Catalog 2019(a)  &Catalog 2019(b)\\ \hline
        Lee et al.(2018)  &92.300  &98.200  &62.472  &95.72  &76.197\\
        MCNN & 92.900  &99.005  &61.897  &96.236  &76.570\\
        KCNN &92.940  &99.203  &62.866  &95.784  &77.402\\
        KMCNN  &93.243  &99.401  &63.302  &96.500  &77.56\\ \hline
      \end{tabular}
      \label{f1}
\end{table}

\begin{table}[h!]
\caption{Precision of KMCNN and other methods}
      \begin{tabular}{cccccc}
        \hline
          &UniProtKB/ & NHGRI-EBI   &MycoSet  &GWAS   &GWAS \\ 
          &Swiss-Prot & GWAS Catalog  & &Catalog 2019(a)  &Catalog 2019(b)\\ \hline
        Lee et al.(2018)  &91.300  &97.300  &61.274  &96.57  &63.452\\
        MCNN & 91.600  &98.418  &60.260  &96.400 &64.390\\
        KCNN &91.613 &98.809  &62.078 &96.043  &65.142\\
        KMCNN  &91.820  &99.203  &61.236  &97.227  &65.283\\ \hline
      \end{tabular}
      \label{precision}
\end{table}

\begin{table}[h!]
\caption{Recall of KMCNN and other methods}
      \begin{tabular}{cccccc}
        \hline
          &UniProtKB/ & NHGRI-EBI   &MycoSet  &GWAS   &GWAS \\ 
          &Swiss-Prot & GWAS Catalog  & &Catalog 2019(a)  &Catalog 2019(b)\\ \hline
        Lee et al.(2018)  &93.400  &99.100  &63.728  &94.812  &95.348\\
        MCNN & 94.200  &99.600  &63.730 &96.640 &95.160\\
        KCNN &94.307 &99.600 &63.730 &95.527  &95.347\\
        KMCNN  &94.710  &99.600  &65.596  &96.704  &95.527\\ \hline
      \end{tabular}
      \label{recall}
\end{table}

\section*{Discussion}
Although it is demonstrated that the knowledge-enhanced CNN outperforms the model without knowledge information, it may cause information redundancy in our experiment, because the same knowledge embedding for each word exists in both channels. In the future work, we would consider to make the knowledge embedding as another independent channel instead of concatenating it with textual features.

From the results shown above, we can tell that KMCNN has the best performance in terms of precision, recall and F1 score over almost all of the datasets. It is demonstrated that both of our ablation models (MCNN and KCNN) outperform the plain CNN adopted by Lee et al. This proves that the information brought by multi-channel and knowledge embeddings could enhance the classification performance.  However, the relative importance of multiple channels and knowledge embeddings is not fixed.

The performance of different models on GWAS Catalog 2019(b) is much lower than that on GWAS Catalog 2019(a) because the negative samples in the former are more confusing. The negative samples are easy to distinguish when they are sampled randomly, which means most of them may have no relationship with genomic variation in their content at all. However, when it comes to publications closely related to genes and diseases simultaneously, classifiers tend to be confused. Further research should be conducted to improve model performance under such circumstances.

\section*{Conclusions}
In this paper, we propose KMCNN model which not only uses multiple channels but also fuses text features and medical knowledge to improve biomedical literature triage. We also develop datasets using asynchronous split so that the model is able to achieve better generalization on future medical literature. To use knowledge to enhance the model’s performance,UMLS2Vec is proposed to generate the embedding of concepts in UMLS.



\begin{backmatter}
\section*{Declaration}
\subsection*{Abbreviations}
KMCNN:  Knowledge-enhanced Multi-channel CNN;  CNN: Convolutional Neural Networks;  UMLS:Unified Medical Language System; GWAS: Genome-wide association study; MCNN: multi-channel CNN; KCNN: knowledge-enhanced CNN

\subsection*{Ethics approval and consent to participate}
Not applicable.

\subsection*{Consent for publication}
Not applicable.

\subsection*{Availability of data and materials}
The datasets generated and/or analysed during the current study are available in the KMCNN repository, https://github.com/wds-seu/KMCNN.

\subsection*{Competing interests}
The authors declare that they have no competing interests.
  
\subsection*{Funding}
The publication of this paper is supported by the National Natural Science Foundation of China under grant U1736204, and the National Key Research and Development Program of China under grant 2018YFC0830201, 2017YFB1002801. This paper is partially funded by the Judicial Big Data Research Center, School of Law at Southeast University.
  
\subsection*{Authors' contributions}
XZ designed the overall study and directed the research project. CL performed the experiments, interpreted and analyzed the results. QL prepared the datasets and drafted the manuscript. XZ and QL reviewed and revised the manuscript. All authors read and approved the final version of the manuscript.

\subsection*{Acknowledgements}
The authors would like to thank the reviewers from the China Conference on Health Information Processing 2019 for their comments, which were incorporated into the present work. We would also like to thank Mr. Da Tong and Mr. Weijian Ye for their valuable suggestions on our work.
  

\bibliographystyle{bmc-mathphys} 
\bibliography{bmc_article}      









\end{backmatter}
\end{document}